\pdfoutput=1

\documentclass[11pt]{article}

\usepackage[preprint]{acl}

\usepackage{times}
\usepackage{latexsym}

\usepackage{tcolorbox}
\tcbuselibrary{listingsutf8} %
\usepackage{listings}
\usepackage{courier}
\usepackage{caption}

\usepackage[T1]{fontenc}

\usepackage[utf8]{inputenc}
\usepackage{pgf-pie}

\usepackage{microtype}

\usepackage{inconsolata}

\usepackage{graphicx}

\title{DocIE@XLLM25: In-Context Learning for Information Extraction using Fully Synthetic Demonstrations}

\author{Nicholas Popovič \hspace{1em} Ashish Kangen \hspace{1em} Tim Schopf \hspace{1em} Michael Färber\\
ScaDS.AI \& TU Dresden, Germany\\
\small\texttt{\{nicholas.popovic,ashish\_yashwanth.kangen,tim.schopf,michael.faerber\}@tu-dresden.de}}

\begin{document}
\maketitle
\begin{abstract}
Large, high-quality annotated corpora remain scarce in document-level entity and relation extraction in zero-shot or few-shot settings.
In this paper, we present a fully automatic, LLM-based pipeline for synthetic data generation and in-context learning for document-level entity and relation extraction.
In contrast to existing approaches that rely on manually annotated demonstrations or direct zero-shot inference, our method combines synthetic data generation with retrieval-based in-context learning, using a reasoning-optimized language model.
This allows us to build a high-quality demonstration database without manual annotation and to dynamically retrieve relevant examples at inference time.
Based on our approach we produce a synthetic dataset of over $5k$ Wikipedia abstracts with approximately $59k$ entities and $30k$ relation triples.
Finally, we evaluate in-context learning performance on the DocIE shared task, extracting entities and relations from long documents in a zero-shot setting.
We find that in-context joint entity and relation extraction at document-level remains a challenging task, even for state-of-the-art large language models.
\end{abstract}

\section{Introduction}

Information extraction (IE) is a key task in natural language processing (NLP) research.
While significant progress has been made in terms of NLP and IE benchmarks in general, few-shot and long context tasks remain relevant and challenging \cite{tan-etal-2022-revisiting,popovic-farber-2022-shot,gui-etal-2024-iepile,xue-etal-2024-autore,diazgarcia2024surveycuttingedgerelationextraction,app15031045}, even in the age of large language models (LLMs).
In this work, we explore this direction in the context of the Document‐level Information Extraction (DocIE) shared task \cite{XLLMACL2025}, which challenges systems to perform joint entity and relation extraction over long, unstructured documents in a zero‐shot setting.\\

Specifically, we approach the topic via two recently popular approaches, LLMs which have been optimized for reasoning-heavy tasks \cite{deepseekai2025deepseekr1incentivizingreasoningcapability}, as well as synthetic data augmentation, which has become popular for IE in particular \cite{li-etal-2023-semi,josifoski-etal-2023-exploiting,DBLP:conf/kbclm/RogulskyP024} as scalable data annotation still represents a major challenge.
In order to combine the two, we construct a simple, retrieval-based in-context learning setting in which an LLM is tasked with extracting entities and relations from a text based on a single example demonstration retrieved based on its similarity to the given text.
In order to preserve the zero-shot setting, we add the constraint that the example demonstration must not be manually annotated, but instead is a synthetically generated example.
We therefore develop an approach for a synthetic data generation pipeline which produces high quality annotated examples of schema-constrained entity and relation extraction.
Our evaluations on the shared task, as well as the Re-DocRED \cite{tan-etal-2022-revisiting} dataset show that in-context joint entity and relation extraction at the document-level remains a challenging task, even for state-of-the-art LLMs.\\

Our contributions in this paper are the following:
\begin{itemize}
    \item We propose a fully automatic pipeline for synthetic data generation based on LLMs.
    \item We apply our pipeline to Wikipedia abstracts and produce a dataset of roughly $5k$ documents annotated with approximately $59k$ entities and $30k$ triples, which we make available for future research\footnote{The code and synthetic dataset are made available at \url{https://github.com/nicpopovic/docie-xllm25}.}.
    \item We present the evaluation of an in-context pipeline which relies on our synthetic demonstrations for in-context learning on the DocIE shared task \cite{XLLMACL2025}.
\end{itemize}

\section{Task Description}

The DocIE shared task \cite{XLLMACL2025} evaluates long form information extraction, specifically entity and relation extraction, in a zero-shot schema-constrained setting:
At test time, a system is provided with only a text document and a schema consisting of strings naming the entity and relation types which are to be extracted.
As is typical for few-shot and zero-shot tasks, the supplied training and development data resembles the test data only in structure. 
It differs in terms of schemata and text domains.\\

\section{Approach}
For this paper, we are interested specifically in putting several key technologies to the test in a pipeline, namely retrieval-based in-context learning, synthetic data \cite{li-etal-2023-semi,josifoski-etal-2023-exploiting,DBLP:conf/kbclm/RogulskyP024}, and reasoning language models \cite{deepseekai2025deepseekr1incentivizingreasoningcapability}. Further, we apply our pipeline to the zero-shot setting using none of the provided training data and foregoing any model fine-tuning.\\

The core idea behind our approach is to construct a fully synthetic demonstration database, from which, given a query, we retrieve the most relevant example and provide it to a reasoning language model as an in-context example.
Below, we first describe the inference pipeline, as it could also be applied to manually annotated data, before describing the synthetic data generation pipeline in detail in Section \ref{sec:synthetic}.

\subsection{Inference Pipeline}

For inference, and thus the evaluation of the DocIE shared task, we use an in-context learning setting split into two LLM calls\footnote{All experiments are performed using DeepSeek-R1-Distill-Qwen2.5-32B \cite{deepseekai2025deepseekr1incentivizingreasoningcapability} at temperature $0$.}.
For the in-context examples, we use a single example document from our synthetic demonstration database, retrieved based on similarity to the query document.
Both calls use the same prompt structure, given in Appendix \ref{sec:appendix-prompts}, Figure \ref{fig:llm-prompt-eval}.
In the first call, we query the model with just the first paragraph of the query document, while the second LLM call supplies the entire document with the annotations provided for the beginning paragraph.
We find that this strategy drastically decreases failures of the model to adhere to the annotation format for long documents.\\

\section{Synthetic Data Generation}
\label{sec:synthetic}
\begin{figure*}
    \centering
    \includegraphics[width=1\textwidth, angle=0]{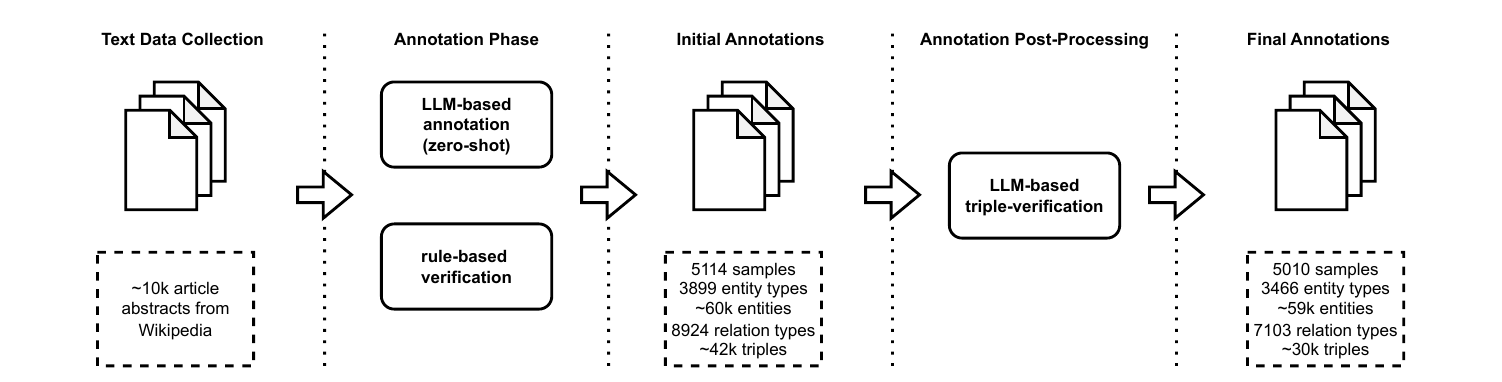}
    \caption{Overview of the pipeline used for synthetic data generation.}
    \label{fig:synth}
\end{figure*}

Figure \ref{fig:synth} provides an overview of the synthetic data generation pipeline which requires only text data as its input and produces complete, high quality annotations over a two-phase process.
\subsection{Text Data Collection}
As a basis for the synthetic dataset we use Wikipedia's Vital Articles (Level 4)\footnote{\url{https://en.wikipedia.org/wiki/Wikipedia:Vital_articles/Level/4}}, which is a collection of 10k articles deemed essential based on criteria such as coverage, notability, and impact on other Wikipedia content. 
These articles represent key topics across various domains and offer a broad scope of vital knowledge.
Detailed information about the category distributions can be found in Appendix \ref{sec:vital_articles_categories}, Figure \ref{fig:vital_articles_categories}.
We create our final synthetic dataset from abstracts which we truncate as a result of initial experiments outlined in section \ref{sec:length_experiments}.
\subsection{Annotation Phase}
The main annotation phase consists of an initial LLM-based annotation followed by rule-based verification of the returned results.\\

The initial annotation run is performed in a zero-shot setting using a reasoning language model, specifically DeepSeek-R1-Distill-Qwen2.5-32B, and a prompt provided in Appendix \ref{sec:appendix-prompts}, Figure \ref{fig:llm-prompt-zs}. The prompt was developed through multiple iterations of manual engineering to meet the following criteria:
\begin{itemize}
    \item \textbf{Zero-Shot Setting}: No use of DocIE shared task data in the prompt.\footnote{Note that the training data was seen by the team constructing the prompts.}
    
    \item \textbf{Machine-Parseable Output}: Requiring the model to return a JSON object yields reliable results.\footnote{We find that properly escaping the input is crucial to avoid syntax errors.}
    
    \item \textbf{Unified Schema}: We use a single prompt for schema generation, entity extraction, and relation extraction. Defining distinct keys for each step in the JSON object ensures coverage and prevents omissions.
    
    \item \textbf{Full NER}: Includes mention detection (as spans), entity typing, and coreference resolution. Though not required for the task, character spans align with standard IE benchmarks and support future encoder-based fine-tuning. Prompting the model to echo the input with inline HTML/XML-style tags for mentions proves robust and allows for automatic validation.
    
    \item \textbf{Verification Hooks for Relations}: We prompt the model to describe each extracted triple in natural language before emitting the structured triple, enabling verification in the post-processing stage. We also require the model to use IDs to refer to the previously extracted entities, which enables automatic consistency checks (e.g., confirming that the subject and object spans actually appear in the source text).
\end{itemize}

To prioritize annotation quality over quantity and enable post-processing, we implement several verification mechanisms after the initial model output. These steps focus particularly on validating extracted relations and ensuring annotation completeness. First, we verify entities by checking that all provided mentions correctly occur within the input text. Second, we validate extracted triples by enforcing that subject and object references are made via entity IDs; we then cross-check that the names referenced in the triples match the previously annotated entities. This ensures the model does not fabricate tuples based on nonexistent mentions.

\subsection{Annotation Post-Processing}

During the construction of the zero-shot prompt, we identified a common source of errors in the directionality of relations (e.g., swapping the subject and object).
The natural language descriptions generated as part of the verification hooks (as outlined above) help mitigate this issue in two ways: First, in our observations, producing a description already improves the accuracy of the extracted triples. Second, these descriptions enable further verification.\\

For each extracted triple, we prompt the model (using the template shown in Appendix \ref{sec:appendix-prompts}, Figure~\ref{fig:llm-prompt-triples}) to assess whether the structured triple faithfully matches its corresponding natural language description.
If an inconsistency is detected, we discard not only the affected triple but the entire set of relations of that type within the document, to avoid including problematic relation types.
This conservative approach prioritizes annotation quality over quantity, in line with our overall dataset construction philosophy.\\

In addition to the triple filtering, we discard annotations in two cases, first if assigned entity types are identical to the entities themselves and second if all entities in a document have the same type.

\section{Experiments and Results}

\subsection{Effects of Text Length on Zero-Shot Annotation}
\label{sec:length_experiments}
Instead of using full abstracts for the synthetic dataset, we truncate the input texts for two reasons: First, shorter texts reduce inference time per document, enabling the annotation of a more diverse set of articles. Second, initial observations indicated that longer texts led to a higher frequency of verification failures when using our prompt.\\

In order to validate the latter observation, we ran an initial annotation round for all abstracts with a length of up to 300 words and tracked error rates.
The results of this experiment, shown in Appendix \ref{sec:error_rates}, Figure \ref{fig:error_rates}, reveal the following:
\begin{itemize}
    \item Text length and total verification failures are highly correlated.
    \item Syntax errors, making the output unparseable and mismatches between the entity IDs used in the triples and the ones extracted from the text are the most common errors. Failures in the entity extraction step, such as extracted entities not occuring in the text or missing span annotations, are less common.
\end{itemize}

\subsection{Statistics of Synthetic Annotation Database}

Based on the results outlined above, to balance inference efficiency, annotation yield, and text length (especially considering that downstream queries will involve longer documents), we implement the following truncation strategy for the remaining data:
We split each abstract into sentences and iteratively add sentences until the cumulative text length has reached 100 words or more and then omit the remaining sentences.
Additionally, upon failed verification, we rerun the zero-shot prompt once with a temperature of $0.2$.
This results in $5114$ annotated documents, a yield of $52.89\%$ for the initial annotation.
After the post-processing phase, the final dataset consists of $5010$ documents, with approx. $59k$ across $3466$ entity types and approx. $30k$ triples across $7103$ relation types.
The most common entity and relation types are shown in Appendix \ref{sec:appendix_dataset_synth} Figures \ref{fig:entity_types} and \ref{fig:relation_types}.
An example of synthetic data is shown in Appendix \ref{sec:data_sample}, Figure \ref{fig:data_sample}.

\subsection{DocIE Test Set Evaluation}

For evaluation on the test set of the DocIE shared task \cite{XLLMACL2025}, we fetch the most similar document from our synthetic dataset based on the similarity measured between the truncated query text (truncated using the same strategy as used during our dataset construction) and the synthetically annotated documents\footnote{We remove 2 documents from our synthetic dataset which are part of the test set in order to avoid contamination.} using the sentence-transformers \cite{reimers-gurevych-2019-sentence} model all-MiniLM-L6-v2\footnote{\url{https://huggingface.co/sentence-transformers/all-MiniLM-L6-v2}}.\\

\begin{table}[ht]
    \centering
    \begin{tabular}{l|ccc}
        \hline
        \textbf{Task} & \textbf{P (\%)} & \textbf{R (\%)} & \textbf{F1 (\%)} \\
        \hline
        Entity Ident. & 52.36 & 23.94 & 32.86 \\
        Entity Class. & 25.79 & 11.80 & 16.19 \\
        RE (General) & 5.03 & 2.44 & 3.29 \\
        RE (Strict) & 4.61 & 2.23 & 3.01 \\
        \hline
    \end{tabular}
    \caption{Evaluation results on the test dataset of the DocIE shared task for entity identification, entity classification, and relation extraction under general and strict modes. Precision (P), Recall (R), and F1 are reported as percentages.}
    \label{tab:results}
\end{table}

The results, shown in Table \ref{tab:results}, place our approach in the fourth place for the shared task.
We note that unparseable outputs occur in a substantial portion of documents, with only $63.91\%$ of outputs being valid.
While we initially hypothesize that this is due, in large parts, to the long documents increasing the frequency of syntax errors in the produced outputs (as this is the trend we observed in our experiments in Section \ref{sec:length_experiments}) our experiments with shorter documents, outlined in Section \ref{sec:eval_redocred}, do not support this.
Even when factoring unparseable outputs into the results, the scores remain low especially for relation extraction.
This suggests that IE, especially relation extraction remains a challenging task in the age of strong LLMs.

\subsection{Evaluation on Re-DocRED}
\label{sec:eval_redocred}
\begin{table}[ht]
    \centering
    \begin{tabular}{l|ccc}
        \hline
        \textbf{Task} & \textbf{P (\%)} & \textbf{R (\%)} & \textbf{F1 (\%)} \\
        \hline
        Mention det. & 72.20 & 25.91 & 38.13 \\
        Entity Ident. & 58.61 & 23.67 & 33.72 \\
        Entity Class. & 52.25 & 21.10 & 30.06 \\
        RE (General) & 13.79 & 1.67 & 2.98 \\
        RE (Strict) & 8.55 & 1.04 & 1.85 \\
        \hline
    \end{tabular}
    \caption{Evaluation results on the test dataset of Re-DocRED \cite{tan-etal-2022-revisiting}. Since we have access to the ground truth labels for this data, we are able to also calculate the mention detection scores.}
    \label{tab:results_redocred}
\end{table}

In Table \ref{tab:results_redocred} we show the results obtained on the test set of the Re-DocRED dataset \cite{tan-etal-2022-revisiting}.
Compared to the DocIE dataset, fewer entity types are used for annotation (person, organization, location, time, number, and miscellaneous), and the documents tend to be shorter.\\

For the overall results, with the exception of entity classification\footnote{We attribute this to the smaller number and more generic kind of entity types used in Re-DocRED.}, the results are broadly comparable to those reported in the DocIE shared task. This suggests that joint entity and relation extraction at the document level remains a challenging problem for current state-of-the-art large language models, even when applied to relatively short documents.\\

\begin{table}[ht]
    \centering
    \begin{tabular}{l|ccc}
        \hline
        \textbf{Task} & \textbf{P (\%)} & \textbf{R (\%)} & \textbf{F1 (\%)} \\
        \hline
        Mention det. & 72.20 & 72.62 & 72.41 \\
        Entity Ident. & 58.61 & 63.73 & 61.06 \\
        Entity Class. & 52.25 & 56.82 & 54.44 \\
        RE (General) & 13.79 & 4.57 & 6.87 \\
        RE (Strict) & 8.55 & 2.84 & 4.26 \\
    \hline
    \end{tabular}
    \caption{Evaluation results on the test dataset of Re-DocRED \cite{tan-etal-2022-revisiting}, restricted to only those documents where the pipeline produced a valid output.}
    \label{tab:results_redocred_valid}
\end{table}

Even though the documents are shorter, with only $38.8\%$ of outputs being valid, producing parseable outputs remains a major challenge.
In order to assess the potential gain of addressing this issue, we measure the results using only those documents where our pipeline produced a valid output.
The corresponding results are shown in Table \ref{tab:results_redocred_valid}.

\section{Conclusion}
We present a fully automatic, LLM-driven pipeline for high-quality synthetic data generation for document-level entity and relation extraction, and apply it to create a dataset of over $5k$ Wikipedia abstracts with roughly $59k$ entities and $30k$ relation triples. Through our two-phase annotation process, zero-shot prompt-based extraction followed by rule- and model-based verification, we demonstrate that automated checks substantially improve annotation fidelity, yielding a dataset that, exhibits consistency and wide coverage of entity and relation types.\\

Our evaluations on the DocIE shared task and Re-DocRED confirm that zero-shot IE remains a hard problem: precision and recall for both entities and relations are modest, and a large number of model outputs still fail to parse cleanly. Moreover, ambiguities in type definitions and relation directionality is a challenge for both extraction and evaluation.\\

By releasing our synthetic dataset we aim to provide a valuable resource for future research on few-shot and zero-shot document-level IE. We hope that these tools will catalyze further advances in robust, scalable information extraction methods.

\section*{Acknowledgements}

This work was funded by the Federal Ministry of Education and Research of Germany and the Saxon State Ministry for Science, Culture and Tourism as part of the Center for Scalable Data Analytics and Artificial Intelligence Dresden/Leipzig (ScaDS.AI, project ID: SCADS24B).
The authors also acknowledge the computing resources provided by the high-performance computing center at the NHR Center of TU Dresden, which is supported jointly by the Federal Ministry of Education and Research and the participating state governments within the NHR framework.

\bibliography{custom, anthology}

\appendix

\section{Prompts}
\label{sec:appendix-prompts}
Figure \ref{fig:llm-prompt-zs} shows the full zero-shot annotation prompt.  
Figure \ref{fig:llm-prompt-triples} shows the triple verification prompt for post-processing extracted relations.  
Figure \ref{fig:llm-prompt-eval} shows the inference prompt (with in-context demonstration) used at test time.

\begin{figure*}[ht]
\centering
\begin{tcolorbox}[title=Zero-Shot Annotation Prompt (DeepSeek-R1-Distill-Qwen-32B), fonttitle=\bfseries, colback=gray!10, colframe=black, width=0.95\linewidth, listing only, listing options={
    basicstyle=\ttfamily\small,
    breaklines=true,
    showstringspaces=false
}]
\begin{verbatim}
Help me build a knowledge graph schema. I will provide a text and you tell me
which entity types and which relation types (properties) to add to my knowledge
graph schema to model the data in the text.
This is the text in question:

{text}

Return your answer in the following format:
```json
{
  'text_with_spans': # html annotated text where every mention and coreference
  of an entity is annotated, for example: '<ent id="0" type="Person">Alice
  </ent> (or <ent id="0" type="Person">Ali</ent> as her friends call her) knows
  <ent id="1" type="Person">Bob</ent> because <ent id="0" type="Person">she
  </ent> met <ent id="1" type="Person">him</ent> at
  <ent id="2" type="Educational institution">school</ent>.',
  'entities': [
    {'id': 0, 'name': <name_of_entity>, 'type': <type_of_entity>},
    ..., # add all entities with the types above, even if they are not relevant
    for a triple
  ],
  'triples': [
    {'description': <text_describing_triple>, 'triple_string':
    '(<name_of_subject>, <name_of_relation_type>, <name_of_object>)',
    'subject': <id_of_subject_entity>, 'predicate': <name_of_relation_type>,
    'object': <id_of_object_entity>},
    ...,
  ],
  'relation_types': [<name_of_relation_type>, ...],
  'entity_types': [<name_of_entity_type>, ...],
}
```

Make sure that for every entity type and relation type you annotate *all*
occurrences!
\end{verbatim}
\end{tcolorbox}
\caption{Prompt used for zero-shot text annotation.}
\label{fig:llm-prompt-zs}
\end{figure*}

\begin{figure*}[ht]
\centering
\begin{tcolorbox}[title=Triple Verification Prompt (DeepSeek-R1-Distill-Qwen-32B), fonttitle=\bfseries, colback=gray!10, colframe=black, width=0.95\linewidth, listing only, listing options={
    basicstyle=\ttfamily\small,
    breaklines=true,
    showstringspaces=false
}]
\begin{verbatim}
Which of the following is a good description of the meaning of the sentence
"{description}"?

A:
```json
{{"subject": "{subject}", "predicate": "{predicate}", "object": "{object}"}}
```
B:
```json
{{"subject": "{object}", "predicate": "{predicate}", "object": "{subject}"}}
```
C:
Both. Only use this option if the predicate/property is a symmetric one.

D:
None of the above. Only use this option if the above are nonsensical or vastly
different from the text.

format your answer like so:
\boxed{{<A_or_B_or_C_or_D>}}
\end{verbatim}
\end{tcolorbox}
\caption{Prompt used for post-processing of triples.}
\label{fig:llm-prompt-triples}
\end{figure*}

\begin{figure*}[ht]
\centering
\begin{tcolorbox}[title=Inference Prompt (DeepSeek-R1-Distill-Qwen-32B), fonttitle=\bfseries, colback=gray!10, colframe=black, width=0.95\linewidth, listing only, listing options={
    basicstyle=\ttfamily\small,
    breaklines=true,
    showstringspaces=false
}]
\begin{verbatim}
Help me build a knowledge graph. I will provide a text and you annotate it.
Here is what correct annotation looks like:
```json
{
    'text': '...',
    'entity_types': [...],
    'text_with_spans': '...',
    'entities': [...],
    'relation_types' + ": [...],
    'relations': [...]
}
```

(Note how the entity ids start from 0 and allow for coreference resolution,
as multiple spans in the annotated text can refer to the same entity.)

Here is the annotation I want you to complete:
```json
{
    'text': '...',
    'entity_types': [...],
    'text_with_spans': '...',
    'entities': [...],
    'relation_types' + ": [...],
    'relations': [...]
}
```

Do not add any entity or relation types! Use only the ones provided in the JSON.
Where possible, reuse the entity ids from the annotation I've started.
If I've missed any entities (or failed to resolve coreferences) or triples,
please fix accordingly.
Return the completed JSON, not just your changes.
\end{verbatim}
\end{tcolorbox}
\caption{Prompt used for inference. An example of an in-context demonstration is shown in Appendix \ref{sec:data_sample}, Figure \ref{fig:data_sample}.}
\label{fig:llm-prompt-eval}
\end{figure*}

\section{Category distribution of Wikipedia Vital Articles}
\label{sec:vital_articles_categories}
\begin{figure*}
\begin{tikzpicture}
  \pie[text=legend, radius=3, explode=0.1, text=pin]{
    19.92/People,
    6.93/History,
    11.98/Geography,
    6.79/Arts,
    4.32/Philosophy and religion,
    4.66/Everyday life,
    9.26/Society and social sciences,
    14.80/Biology and health sciences,
    10.98/Physical sciences,
    7.37/Technology,
    2.98/Mathematics
  }
\end{tikzpicture}
\caption{Category distribution of level 4 vital articles (approx. 10k).} 
\label{fig:vital_articles_categories}
\end{figure*}
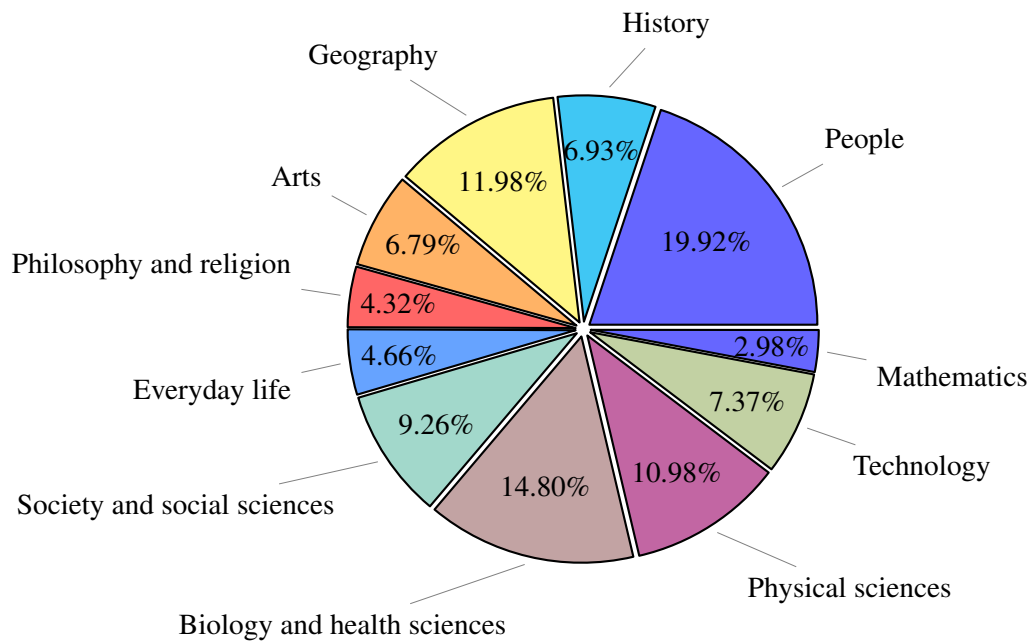

Figure \ref{fig:vital_articles_categories} shows the pie-chart distribution of Level 4 Vital Articles by domain category.

\section{Effects of Text Length on Errors in Zero-Shot Annotation}
\label{sec:error_rates}
\begin{figure*}
    \centering
    \includegraphics[width=1\textwidth, angle=0]{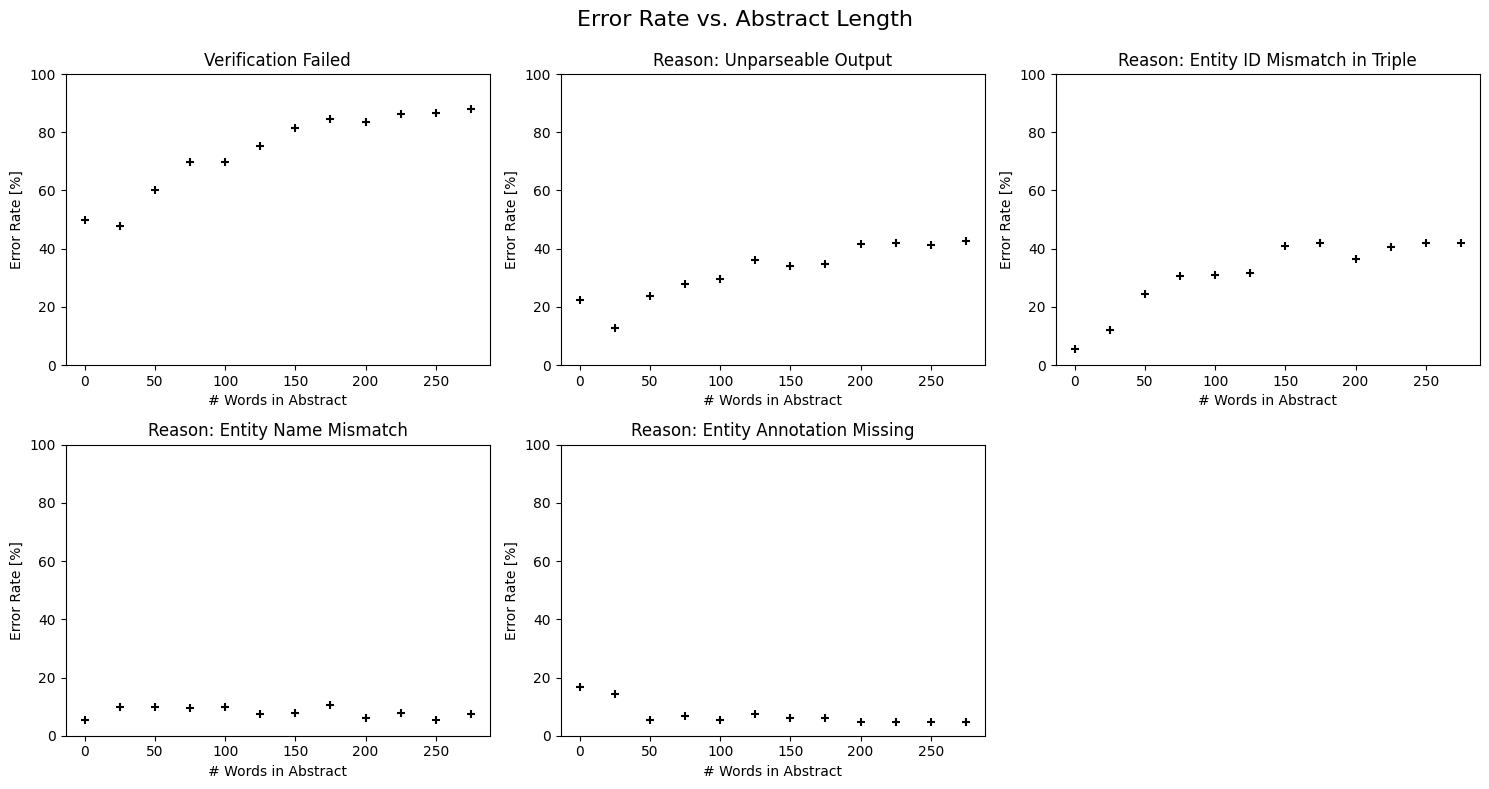}
    \caption{Error distribution for initial set of experiments of first annotation phase.}
    \label{fig:error_rates}
\end{figure*}

Figure \ref{fig:error_rates} shows how error rates (syntax failures, span mismatches, etc.) vary with input text length.

\section{Most Common Entity and Relation Types}
\label{sec:appendix_dataset_synth}
\begin{figure*}
    \centering
    \includegraphics[width=1\textwidth, angle=0]{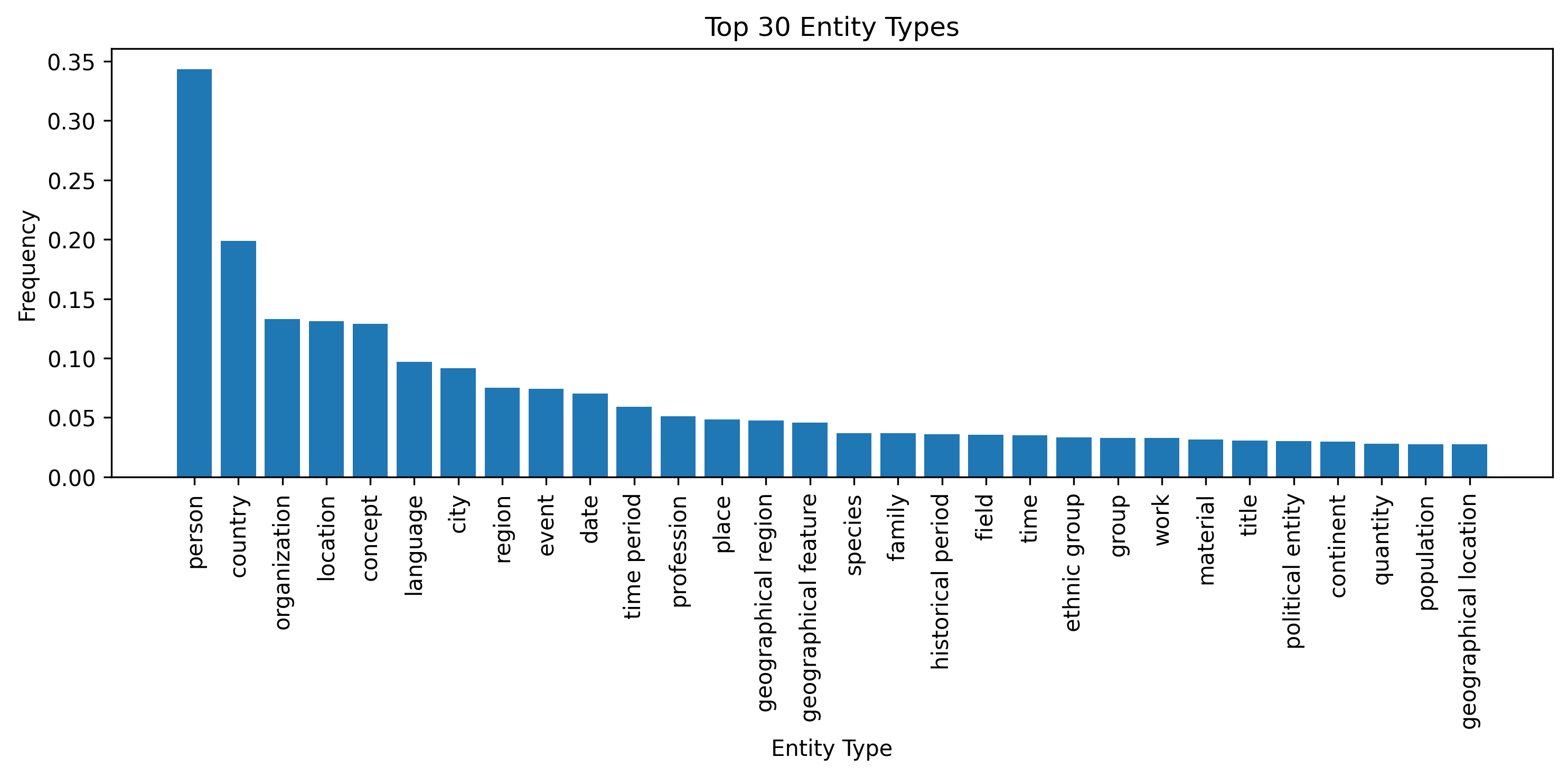}
    \caption{Top 30 entity types in final synthetic dataset and their frequency.}
    \label{fig:entity_types}
\end{figure*}

\begin{figure*}
    \centering
    \includegraphics[width=1\textwidth, angle=0]{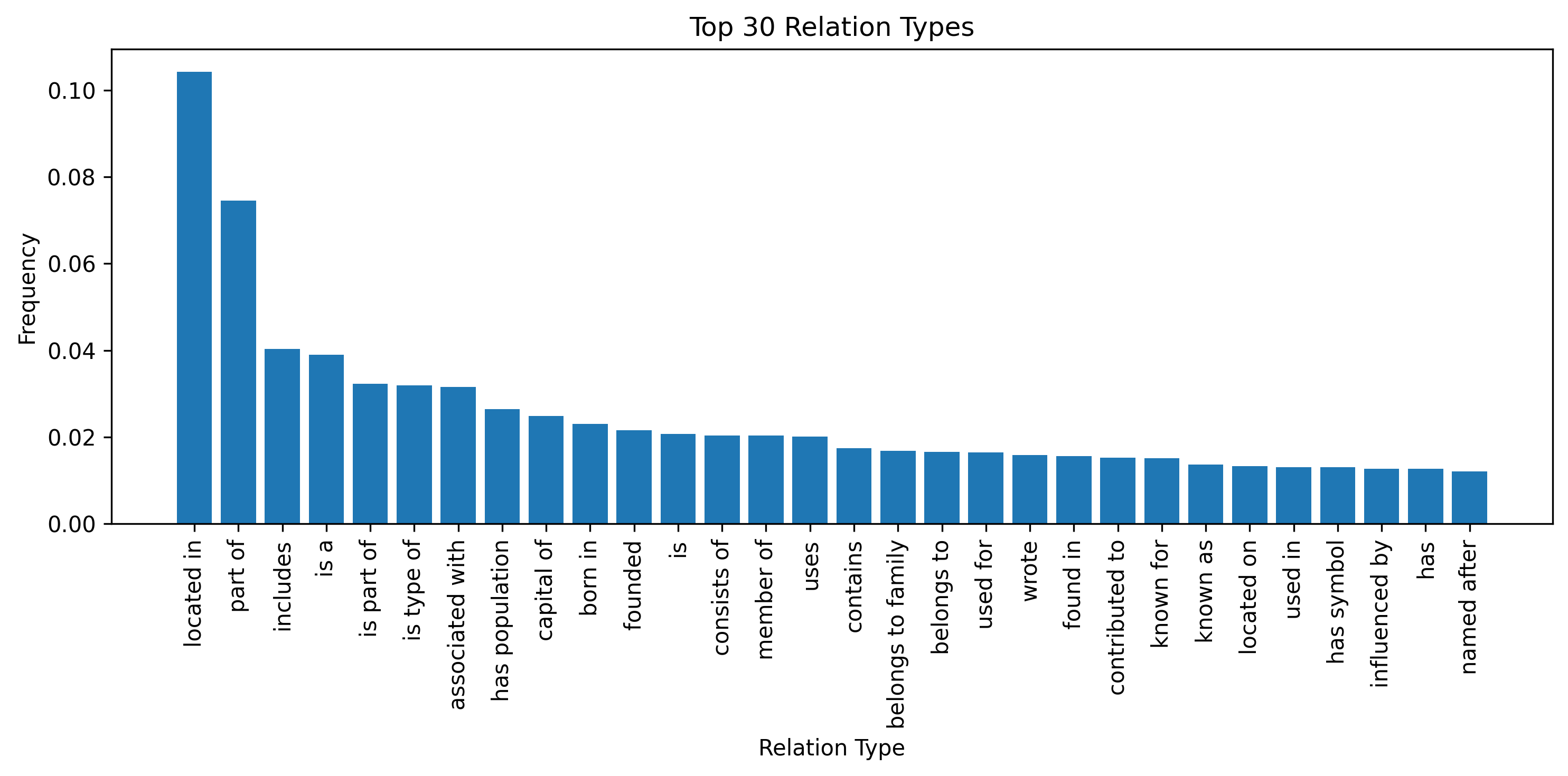}
    \caption{Top 30 relation types in final synthetic dataset and their frequency.}
    \label{fig:relation_types}
\end{figure*}

Figure \ref{fig:entity_types} shows the top 30 entity types and their frequencies in the synthetic dataset.  
Figure \ref{fig:relation_types} shows the top 30 relation types and their frequencies.


\section{Data Sample}
\label{sec:data_sample}
Figure \ref{fig:data_sample} shows an example from the synthetic dataset.

\begin{figure*}[ht]
\centering
\begin{tcolorbox}[title=Synthetic Data Example, fonttitle=\bfseries, colback=gray!10, colframe=black, width=0.95\linewidth, listing only, listing options={
    basicstyle=\ttfamily\small,
    breaklines=true,
    showstringspaces=false
}]
\begin{tiny}
\begin{verbatim}
{
    "text": "The University of Vienna (German: Universit\u00e4t Wien) is a public research university located in Vienna, Austria. Founded by Duke
    Rudolph IV in 1365, it is the oldest university in the German-speaking world and among the largest institutions of higher learning in Europe. 
    The university is associated with 17 Nobel Prize winners and has been the home to many scholars of historical and academic importance.",
    
    "annotated_text": "<ent id=\"0\" type=\"Educational institution\">The University of Vienna</ent> (German: 
    <ent id=\"0\" type=\"Educational institution\">Universit\u00e4t Wien</ent>) is a public research university located in 
    <ent id=\"1\" type=\"Location\">Vienna</ent>, <ent id=\"2\" type=\"Country\">Austria</ent>. Founded by 
    <ent id=\"3\" type=\"Person\">Duke Rudolph IV</ent> in 1365, it is the oldest university in the German-speaking world and among the largest 
    institutions of higher learning in Europe. The university is associated with 17 <ent id=\"4\" type=\"Award\">Nobel Prize</ent> winners and
    has been the home to many <ent id=\"5\" type=\"Person\">scholars</ent> of historical and academic importance.",
    
    "entities": [
        {
            "id": 0,
            "name": "The University of Vienna",
            "type": "Educational institution"
        },
        {
            "id": 1,
            "name": "Vienna",
            "type": "Location"
        },
        {
            "id": 2,
            "name": "Austria",
            "type": "Country"
        },
        {
            "id": 3,
            "name": "Duke Rudolph IV",
            "type": "Person"
        },
        {
            "id": 4,
            "name": "Nobel Prize",
            "type": "Award"
        },
        {
            "id": 5,
            "name": "scholars",
            "type": "Person"
        }
    ],
    
    "relations": [
        {
            "description": "The University of Vienna is located in Vienna.",
            "triple_string": "(The University of Vienna, located_in, Vienna)",
            "subject": 0,
            "predicate": "located_in",
            "object": 1
        },
        {
            "description": "The University of Vienna is located in Austria.",
            "triple_string": "(The University of Vienna, located_in, Austria)",
            "subject": 0,
            "predicate": "located_in",
            "object": 2
        },
        {
            "description": "The University of Vienna was founded by Duke Rudolph IV.",
            "triple_string": "(The University of Vienna, founded_by, Duke Rudolph IV)",
            "subject": 0,
            "predicate": "founded_by",
            "object": 3
        },
        {
            "description": "The University of Vienna has been the home to scholars.",
            "triple_string": "(The University of Vienna, home_of, scholars)",
            "subject": 0,
            "predicate": "home_of",
            "object": 5
        }
    ]
}
\end{verbatim}
\end{tiny}
\end{tcolorbox}
\caption{Example of final synthetic data.}
\label{fig:data_sample}
\end{figure*}

\end{document}